\documentclass[preprint,12pt]{elsarticle}

\usepackage{amsmath,amssymb,amsfonts}
\usepackage{mathtools}
\usepackage{bm}
\usepackage{siunitx}

\usepackage{booktabs}
\usepackage{array}
\usepackage{tabularx}
\usepackage{multirow}
\usepackage{ragged2e}
\newcolumntype{L}{>{\RaggedRight\arraybackslash}X}
\usepackage{adjustbox}

\usepackage{subcaption}
\usepackage{enumitem}

\usepackage{tikz}
\usetikzlibrary{positioning,arrows.meta,fit,shapes.geometric,calc}

\usepackage[hyphens]{url}

\biboptions{sort&compress}

\newcommand{\encoderstudyref}{the companion encoder study~\citep{odeyemi_encoder}}
\newcommand{\encoderstudyshort}{companion-study}
\newcommand{\thisdocnoun}{study}

\newcolumntype{Y}[1]{>{\hsize=#1\hsize\RaggedRight\arraybackslash}X}

\journal{Artificial Intelligence in Medicine}

\begin{document}

\begin{frontmatter}

\title{Intact-to-Amputee Transfer in Surface-EMG Gesture Decoding: Training Source and
Calibration Budget}

\author[bme]{Jethro Odeyemi}
\ead{jethro.odeyemi@usask.ca}

\author[bme]{W.J. (Chris) Zhang\corref{cor1}}
\ead{chris.zhang@usask.ca}
\cortext[cor1]{Corresponding author.}

\affiliation[bme]{organization={Division of Biomedical Engineering, University of Saskatchewan},
            addressline={57 Campus Drive},
            city={Saskatoon},
            state={SK},
            postcode={S7N 5A9},
            country={Canada}}

\begin{abstract}
A recogniser trained on one person rarely transfers to the next, and useful performance usually
demands a fresh round of labelled calibration from the end user. A systematic review of 1077
studies quantifies where the evidence is thin: amputees appear in about one in six. Here a
montage-agnostic cross-user encoder is carried to eleven transradial amputees on a protocol
matched to its intact-limb training data. Zero-shot cross-population transfer fails outright: the
encoder requires labeled data from the new user before it begins decoding, and it then exceeds the
per-user classifier a clinic would fit by 0.190 macro F1 at three repetitions and for every
subject in the cohort. Given three labelled repetitions it reaches 0.779 macro-F1 against 0.589
for the per-user pipeline. Training on forty intact subjects produces better transfers to a new
amputee than training on ten other amputees, and combining the two produces better transfers than
either individually. The prediction pre-registered for this study, which extends the encoder's
baseline-strength account with the premise that amputee EMG is less separable, holds true only
after a few repetitions become available and after enriching the source pool with additional
amputees. At a single repetition, and at every budget under a source matched to the intact-limb
comparison, it fails. Thus, it locates the boundary of the proposed account.
\end{abstract}

\begin{keyword}
surface electromyography \sep upper-limb prosthetics \sep transradial amputation \sep transfer
learning \sep gesture recognition \sep calibration budget
\end{keyword}

\end{frontmatter}

\section{Introduction}

The systematic review conducted for this series~\citep{odeyemi_review} shows that in 162 of 1,077
studies (about one in six), researchers studied amputees. In contrast, studies of able-bodied
individuals totaled 656. Researchers build their methodologies, adjust their parameters, and
report accuracy based solely upon a population that does not require the device. Therefore,
although the methodology was developed and validated on a specific population, it has not yet
demonstrated its ability to function within the target population.

In addition to recruitment issues, after a transradial amputation the forearm length will be
shortened. Muscles will be removed, repositioned, or atrophied. Additionally, users will be
generating a movement that they cannot visually confirm or physically sense (i.e.,
proprioception). As such, the electrical signals reaching the electrodes will likely be
weaker/noisier and less easily differentiated than those produced by an intact forearm during a
similar movement. Furthermore, an amputee's movement has no physical consequence to help
stabilize it. Therefore, the decoder trained on intact limbs will have learned regularities that
may not be present in the target population.

The purpose of this paper is to quantify this gap and determine what can be done to fill it. To
accomplish this, a matched transfer-pair study design is implemented. Eleven transradial amputees
record movements on the same system (NinaPro DB3) as forty able-bodied participants recorded
movements in DB2. As such, the hardware, acquisition system (12 channels at 2000 Hz), and movement
protocols are identical for both populations. However, pool sizes are not equal. There are only
eleven amputees in the matched dataset and this limitation is addressed in the Discussion section.
For each amputee whose data were left out of the training set for a particular fold, three
different training datasets are evaluated using the same calibration repetitions. The training
datasets included: Forty able-bodied participants, Ten additional amputees, or Both. Each of these
datasets are compared to a per-user classifier that would be fitted for each participant by a
clinical team.

There are three primary findings from this study. First, without any labeled data from the
amputee, transferring failed completely. Second, even when the amputee supplies only a few
labeled calibration repetitions, training on the 40 able-bodied participants resulted in higher
classification performance than training on the 10 other amputees. Thirdly, combining both
groups (both able bodied and amputees) resulted in higher classification performance than either
group alone. Importantly, with both groups combined, all 11 amputees exceeded their respective
per-user baseline classifiers with just three calibration repetitions.

\section{Background and positioning}

Amputee myoelectric decoding research has existed for decades. One consistent finding among
researchers is that the same decoding pipeline achieves poorer performance in amputee
populations than in intact populations~\citep{atzori2014electromyography,atzori2016deep}. This
pattern has also been consistently observed across the amputee
corpus~\citep{atzori2016deep,zhai2017self,phinyomark2018feature,parajuli2019real,krasoulis2017improved,cheesborough2015targeted}.
The NinaPro programme released DB3 so that this could be measured on shared data, pairing it
with the intact DB2 recorded under the same
protocol~\citep{atzori2014ninapro,pizzolato2017comparison}. Reported causes include shorter
residual muscle, atrophy, surgical reorganisation, longer time since amputation, and the absence
of proprioceptive feedback confirming the attempted
movement~\citep{atzori2016deep,zhai2017self,phinyomark2018feature}.

Research into closing this gap includes two primary approaches: transfer learning/domain
adaptation; and architectural improvements to reduce degradation when processing low-quality
signals. Transfer learning/domain adaptation uses adversarial and statistical alignment
techniques reviewed in a companion
study~\citep{odeyemi_intersession,ganin2016domain,du2017intersession} to leverage abundant
data for training followed by adaptation to sparse data. Architectural modifications argue that
models using multiple electrodes with shared weights degrade less rapidly when some of those
electrodes contribute less signal. However, the literature infrequently compares two distinct
questions: Does intact-limb data improve an amputee's classification performance? Is intact-limb
data more effective than existing limited amounts of data from the amputee population?

This paper uses the same encoder architecture as the companion
studies~\citep{odeyemi_encoder,odeyemi_intersession,odeyemi_fewshot}. The encoder
architecture is constructed to be montage-agnostic. The architecture was designed to read the
number of active electrodes directly from the input data instead of relying on a fixed
assumption regarding how many electrodes were being used. Amputees who utilize fewer functional
electrodes would not need to undergo resampling or manually add unused channels or train
separate models. Because none of the eleven subjects in DB3 utilized fewer than twelve
electrodes (Table~\ref{tbl:ch6-cohort}), this benchmark does not assess that capability.

\section{Method}

\subsection{Montage-agnostic encoder architecture}

\begin{figure}[tbp]\centering
\begin{subfigure}[b]{0.88\linewidth}\centering
\resizebox{\linewidth}{!}{%
\begin{tikzpicture}[
  font=\small,
  box/.style={draw, rounded corners, align=center, minimum height=8mm, inner sep=4pt, fill=black!4},
  op/.style={draw, rounded corners, align=center, minimum height=8mm, inner sep=4pt, fill=blue!7},
  hl/.style={draw, rounded corners, align=center, minimum height=8mm, inner sep=4pt, fill=green!8},
  ->, >=Latex, node distance=6mm and 9mm]

\node[box] (in) {Multichannel sEMG\\window $x\in\mathbb{R}^{C\times T}$\\($C\in\{8,10,12,16\}$)};
\node[op, right=of in] (rtn) {Rolling-time\\normalisation\\(causal, per channel)};
\node[op, right=of rtn] (cnn) {Shared per-channel\\temporal CNN\\(weights tied across $C$)};
\node[hl, right=of cnn] (pos) {$+$ electrode-coordinate\\position encoding\\$\mathrm{MLP}(x_c,y_c,z_c)$};

\node[op, below=17mm of pos] (acm) {Channel\\masking (train only,\\bounded, keep $\ge 3$)};
\node[op, left=of acm] (tf) {Cross-channel\\Transformer\\($N$ layers, attention\\over $C$ tokens)};
\node[op, left=of tf] (pool) {Attention\\pooling over\\channel tokens};
\node[box, left=of pool] (out) {Window\\embedding\\$z\in\mathbb{R}^{d}$};

\draw (in) -- (rtn);
\draw (rtn) -- (cnn);
\draw (cnn) -- (pos);
\draw (pos.south) -- (acm.north);
\draw (acm) -- (tf);
\draw (tf) -- (pool);
\draw (pool) -- (out);

\node[align=center, font=\footnotesize, above=3mm of cnn] {\emph{one token per electrode}};
\node[align=center, font=\footnotesize, fill=white, inner sep=1pt, below=3mm of tf]
  {\emph{position from geometry, not channel index}};

\node[hl, below=17mm of out] (head) {Linear gesture head};
\node[hl, right=7mm of head] (mae) {Envelope-reconstruction\\head (pretraining)};
\draw (out.south) -- (head.north);
\draw[dashed] (out.south) -| (mae.north);
\end{tikzpicture}%
}
\caption{}\label{fig:ch3-encoder}
\end{subfigure}

\vspace{2ex}

\begin{subfigure}[b]{0.94\linewidth}\centering
\adjustbox{max width=\linewidth}{%
\begin{tikzpicture}[
  font=\small,
  b/.style={rounded corners=2pt, draw, minimum height=8mm, inner xsep=6pt, align=center},
  src/.style={b, fill=blue!8, draw=blue!55},
  amp/.style={b, fill=orange!12, draw=orange!60},
  tgt/.style={b, fill=red!8, draw=red!55},
  mdl/.style={b, fill=green!10, draw=green!55},
  arr/.style={-{Latex[length=2mm]}, thick},
  lbl/.style={font=\footnotesize\itshape, text=black!60}]

\node[src, minimum width=34mm] (db2) at (0,2.4) {\textbf{DB2}: 40 intact subjects\\ \footnotesize 12 electrodes, 2\,kHz};
\node[amp, minimum width=34mm] (others) at (0,0.9) {\textbf{DB3}: the other 10 amputees\\ \footnotesize same rig, same protocol};
\node[tgt, minimum width=34mm] (held) at (0,-0.9) {\textbf{held-out amputee}\\ \footnotesize $k$ calibration reps $\mid$ test reps};

\node[mdl, minimum width=30mm] (c1) at (5.6,3.0) {intact only};
\node[mdl, minimum width=30mm] (c2) at (5.6,1.7) {amputee pooled};
\node[mdl, minimum width=30mm] (c3) at (5.6,0.4) {intact + amputee};
\node[b, fill=black!5, minimum width=30mm] (c4) at (5.6,-1.4) {per-user LDA\\ \footnotesize comparison baseline};

\draw[arr, blue!55] (db2.east) .. controls +(1.2,0) and +(-1.2,0) .. (c1.west);
\draw[arr, blue!55] (db2.east) .. controls +(1.2,0) and +(-1.2,0) .. (c3.west);
\draw[arr, orange!65] (others.east) .. controls +(1.2,0) and +(-1.2,0) .. (c2.west);
\draw[arr, orange!65] (others.east) .. controls +(1.2,0) and +(-1.2,0) .. (c3.west);
\draw[arr, red!60] (held.east) .. controls +(1.2,0) and +(-1.2,0) .. (c4.west);

\node[b, fill=black!4, minimum width=26mm] (cal) at (10.2,1.7)
  {calibrate on $k$ reps\\ \footnotesize of the held-out amputee};
\node[b, fill=red!7, draw=red!55, minimum width=26mm] (ev) at (10.2,-0.4)
  {evaluate on that\\ amputee's test reps};
\draw[arr] (c1.east) .. controls +(1.0,0) and +(-1.0,0.4) .. (cal.west);
\draw[arr] (c2.east) -- (cal.west);
\draw[arr] (c3.east) .. controls +(1.0,0) and +(-1.0,-0.4) .. (cal.west);
\draw[arr] (cal.south) -- (ev.north);
\draw[arr] (c4.east) .. controls +(1.6,0) and +(-1.2,-0.3) .. (ev.west);
\draw[arr, red!55, dashed] (held.south) .. controls +(3.0,-1.2) and +(-1.6,-1.2) .. node[lbl,below]{test reps never trained on} (ev.south);

\end{tikzpicture}%
}
\caption{}\label{fig:ch6-design}
\end{subfigure}
\caption{Method. (a) The proposed montage-agnostic encoder. Each electrode
becomes a single token whose position is supplied by its forearm coordinate rather than its channel
index, so one set of weights ingests any electrode count. Rolling-time normalisation makes the
representation calibration-free, cross-channel attention mixes the electrode tokens, and bounded
channel masking regularises toward user-invariant features. The linear head is used for gesture
classification; the dashed head is the self-supervised pretraining branch. (b) The transfer design.
DB2 and DB3 were recorded with the same twelve electrodes at the
same rate under the same movement protocol, so the participant population is the variable under
test. For each held-out amputee three training sources are compared, each calibrated on the same
$k$ labelled repetitions from that subject and scored on that subject's held-out test
repetitions, against a per-user classifier fitted on those same repetitions.}
\label{fig:ch6-method}
\end{figure}
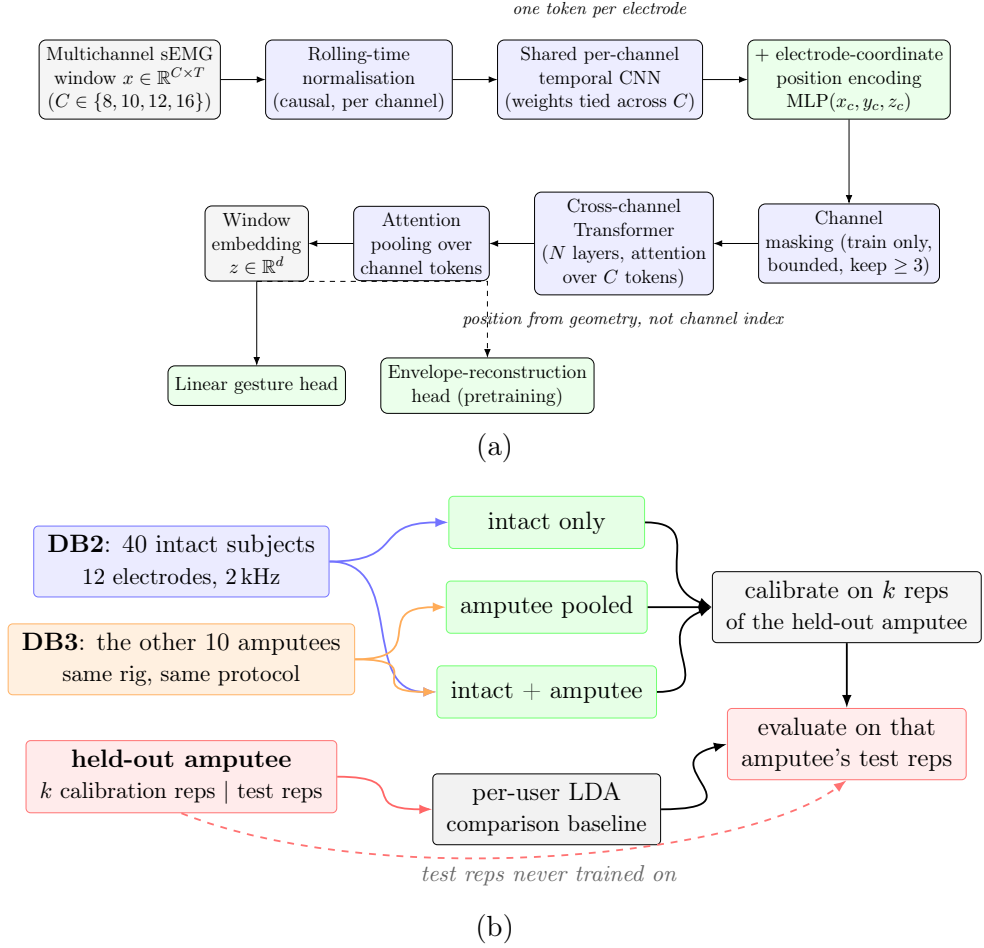

We begin by providing details concerning our encoder's design. The overall architecture is
depicted in Figure~\ref{fig:ch3-encoder}. Our encoder maps sequences of EMG signals ($x \in
\mathbb{R}^{C\times T}$) onto fixed-size embeddings ($z$), where $C$ represents the possible
number of channels in any of the available montages.

Our proposed encoder is founded on the idea that each electrode is treated as one independent
token whose identity corresponds to its physical position on the forearm.

\paragraph{Causal Rolling-Time normalisation.} Our first layer normalizes each channel
individually using their respective causal rolling statistics over time; these statistics
represent expanding window means and variances calculated over time. Since our normalizing is
done per channel and only uses causal information from past time steps, at runtime a previously
unobserved user will self-normalize their input data with respect to their own signal using only
their own causal history; no user-specific statistics will be transported from training to
testing. Hence, even though there is no labeled data from the new user available at runtime, our
model will still adaptively scale its input representations according to new users' signals
without requiring any additional labels.

\paragraph{Shared per-channel tokeniser.} To produce one feature vector per electrode, we apply
a single one-dimensional convolutional stack separately to every channel. Due to sharing of
weights across channels, adding or removing an electrode simply adds/removes a token to/from our
tokenizer without modifying any parameter(s).

\paragraph{Electrode-coordinate position encoding.} Each electrode contains normalized $(x,y,z)$
coordinates specifying its position on the forearm. These coordinates are mapped to vectors
added to each electrode token by means of a multilayer perceptron. Therefore, our model learns
where each electrode resides, not merely its index, thus allowing any electrode configuration to
map into one uniform representation space, with each electrode serving as an unordered element
within that space.

\paragraph{Cross-channel attention.} Our transformer encoder~\citep{vaswani2017attention}
applies attention mechanisms across our electrode tokens. The temporal structure is encoded via
our per-channel tokenizers; attention mechanisms capture spatial relations between electrodes.
Following application of attention mechanisms across all channels, an attention-pooling layer is
used to collapse the electrode tokens into a single window embedding; finally a linear head
computes gesture logits.

\paragraph{Channel-masking regularisation.} During training time, a bounded number of electrodes
is randomly masked such that at least three channels remain active for each montage. This forces
our model towards features that are invariant or robust with regards to missing or displaced
electrodes rather than relying on any single channel. How many electrodes should be masked
affects whether masking fails completely: if too aggressive on very small montages (e.g., all
electrodes masked), there are no valid windows containing active electrodes resulting in
collapsed training; we solved this issue by enforcing bounds for masking.

\subsection{The matched transfer pair}

The matched transfer pair experimental design is illustrated in Figure~\ref{fig:ch6-design}. The
databases are summarized in Table~\ref{tbl:ch6-dataset}. DB2 contains forty intact subjects and
DB3 contains eleven transradial amputees; each group is processed utilizing twelve electrodes
sampled at 2000Hz. Both groups completed six repetitions of the same 49-movement set plus rest.
All processing steps were executed similarly to every other experiment presented within this
series~\citep{odeyemi_encoder,odeyemi_intersession,odeyemi_fewshot}.

Leave-One-Out Amputee Evaluation was applied. For each of the held-out amputee subjects, the
encoder was trained from one of three possible sources and calibrated using $k$ labeled
repetitions for that specific subject. Then, each encoder was tested on that same subject's
unseen testing repetitions; no stages of training or calibration had access to these testing
repetitions. Four budgets ($k = 0$, $k = 1$, $k = 3$, $k = 4$) were assessed; four is the
maximum budget since two of each subject's six repetitions are withheld for testing. Three
sources are detailed in Table~\ref{tbl:ch6-sources}: intact-only; all ten amputees pooled
together; and both intact-only and amputee pooled.

\begin{table}[t]\centering
\caption{The matched transfer pair, and the training configuration for the transfer experiments.
DB2 and DB3 were recorded on the same rig with the same protocol, so the participant population is
the variable under test.}
\label{tbl:ch6-dataset}\label{tbl:ch6-hyperparams}
{\footnotesize\setlength{\tabcolsep}{3pt}%
\adjustbox{max width=\linewidth}{%
\begin{tabular}{lcc}
\toprule
Property & DB2 (source) & DB3 (target) \\
\midrule
Participants & 40 intact & 11 transradial amputees \\
Electrodes & 12 & 12 \\
Sampling rate & 2 kHz & 2 kHz \\
Movements & 49 + rest & 49 + rest (as performed) \\
Repetitions & 6 & 6 \\
Window / stride & 300 / 150 ms & 300 / 150 ms \\
\midrule
Hyper-parameter & \multicolumn{2}{c}{Value} \\
\midrule
Base-training epochs & \multicolumn{2}{c}{40} \\
Calibration epochs & \multicolumn{2}{c}{30} \\
Learning rate & \multicolumn{2}{c}{$4\times10^{-4}$ (calibration $3\times10^{-4}$)} \\
Windows per intact subject & 4000 & -- \\
Windows per amputee subject & -- & 8000 \\
Model width $d$ / layers / heads & \multicolumn{2}{c}{256 / 4 / 4} \\
Channel masking & \multicolumn{2}{c}{bounded, up to 25\%} \\
Hardware & \multicolumn{2}{c}{NVIDIA A100 (3g.20gb), one task per held-out amputee} \\
\bottomrule
\end{tabular}}}
\end{table}

\subsection{The comparison baseline}

The baseline employed is the same Hudgins Feature Set (linear Discriminant Classifier) trained
on only the held-out amputee's own $k$ calibration repetition(s); the same baseline was employed
throughout this series of
studies~\citep{odeyemi_encoder,odeyemi_intersession,odeyemi_fewshot} and represents the type of
pipeline a clinician would develop for an individual patient. When $k=0$ (zero repetitions) it
obviously cannot be trained as it has no information to use; unlike a transferred model that can
continue to operate regardless of whether the user has supplied labels.

\subsection{Cohort composition and per-subject reporting}

Amputees show greater differences amongst themselves compared to the intact subjects. Two of the
eleven did not complete the full set of movements. No class is removed; if a subject never
performs a movement, the subject will have none to evaluate for that movement, and
Table~\ref{tbl:ch6-cohort} shows what each subject contributes. All results are shown per
subject and in the mean.

\section{Experimental setup}

Each of the eleven folds was executed independently on NVIDIA A100 GPUs, with one held-out
amputee per task, using the hyperparameters found in Table~\ref{tbl:ch6-hyperparams}. Because
the source pools differ in terms of their size, a fixed number of windows is sampled from each
subject instead of each condition. Each intact subject is given 4000 windows, whereas each
amputee is provided 8000 windows. This gives 160,000 total windows for the intact condition
versus 80,000 total windows for the amputee condition. As such, the amputee condition draws
twice as heavily on each of its subjects. The remaining imbalance between conditions is due to
the number of subjects per condition and not the number of windows drawn per subject.

\section{Results}

\begin{table}[t]\centering
\caption{What each training source contains, and the full transfer matrix. Intact-to-amputee
transfer on NinaPro DB3, leave-one-amputee-out. Trial-voted macro-F1, mean over the eleven
amputees, for each training source and calibration budget, against a per-user LDA baseline fitted
on the same repetitions. Cells give the macro-F1 mean and across-subject standard deviation for
every training source and budget. Every condition is evaluated on the same held-out amputee and
never sees that subject's test repetitions.}
\label{tbl:ch6-sources}\label{tbl:ch6-headline}\label{tbl:ch6-matrix}
{\scriptsize\setlength{\tabcolsep}{2pt}%
\begin{tabularx}{\textwidth}{@{}l Y{0.86} Y{1.14} cccc@{}}
\toprule
Condition & Training data & Question it answers & 0-shot & 1-shot & 3-shot & 4-shot \\
\midrule
intact only & 40 DB2 intact subjects & does intact-limb data transfer at all? &
  0.018 (0.020) & 0.369 (0.146) & 0.687 (0.198) & 0.747 (0.219) \\
amputee pooled & the other 10 DB3 amputees & is same-population data enough? &
  0.017 (0.017) & 0.285 (0.142) & 0.552 (0.270) & 0.605 (0.281) \\
intact + amputee & both of the above & does combining them help? &
  0.023 (0.025) & 0.464 (0.168) & 0.779 (0.215) & 0.844 (0.211) \\
\midrule
per-user LDA & the held-out amputee's own $k$ reps & the per-user baseline &
  -- & 0.405 & 0.589 & 0.625 \\
\bottomrule
\end{tabularx}}
\end{table}

\begin{table}[t]\centering
\caption{The DB3 amputee cohort as parsed, twelve electrodes for every subject. Two subjects did
not perform the full movement set; no class is dropped, so their missing movements are simply
absent from their own evaluation. The score columns give per-amputee macro-F1 at three calibration
repetitions: the combined-source encoder exceeds the per-user LDA baseline for every subject.}
\label{tbl:ch6-cohort}\label{tbl:ch6-persubject}
{\footnotesize\setlength{\tabcolsep}{4pt}%
\begin{tabular}{lcccccccc}
\toprule
Subject & Windows & \shortstack{Movement\\classes} & \shortstack{Intact\\only} &
  \shortstack{Amputee\\pooled} & \shortstack{Intact +\\amputee} & LDA & Enc.$-$LDA \\
\midrule
s1 & 8361 & 39 & 0.778 & 0.761 & 0.909 & 0.657 & +0.252 \\
s2 & 10668 & 50 & 0.828 & 0.710 & 0.913 & 0.645 & +0.268 \\
s3 & 8248 & 50 & 0.702 & 0.001 & 0.828 & 0.662 & +0.166 \\
s4 & 12773 & 50 & 0.739 & 0.752 & 0.800 & 0.679 & +0.121 \\
s5 & 11528 & 50 & 0.672 & 0.608 & 0.700 & 0.536 & +0.164 \\
s6 & 12642 & 50 & 0.651 & 0.642 & 0.807 & 0.681 & +0.126 \\
s7 & 12501 & 50 & 0.107 & 0.017 & 0.137 & 0.132 & +0.006 \\
s8 & 10471 & 50 & 0.757 & 0.613 & 0.920 & 0.598 & +0.322 \\
s9 & 10007 & 50 & 0.848 & 0.720 & 0.919 & 0.750 & +0.169 \\
s10 & 10146 & 43 & 0.623 & 0.472 & 0.760 & 0.507 & +0.253 \\
s11 & 10092 & 50 & 0.850 & 0.779 & 0.877 & 0.630 & +0.247 \\
\midrule
Total & 117437 & -- & & & & & \\
Mean & & & 0.687 & 0.552 & 0.779 & 0.589 & +0.190 \\
\bottomrule
\end{tabular}}
\end{table}

\subsection{Zero-shot cross-population transfer}

\begin{figure}[tbp]\centering
\begin{subfigure}[b]{0.49\linewidth}\centering
\includegraphics[width=\linewidth]{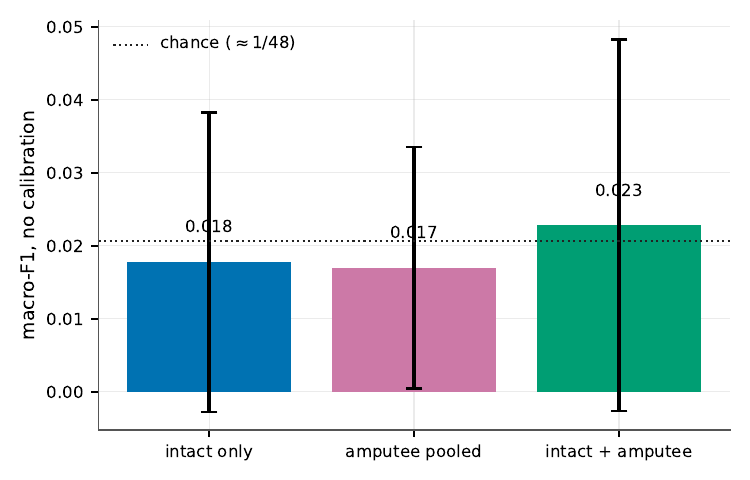}
\caption{}\label{fig:ch6-zeroshot}
\end{subfigure}\hfill
\begin{subfigure}[b]{0.49\linewidth}\centering
\includegraphics[width=\linewidth]{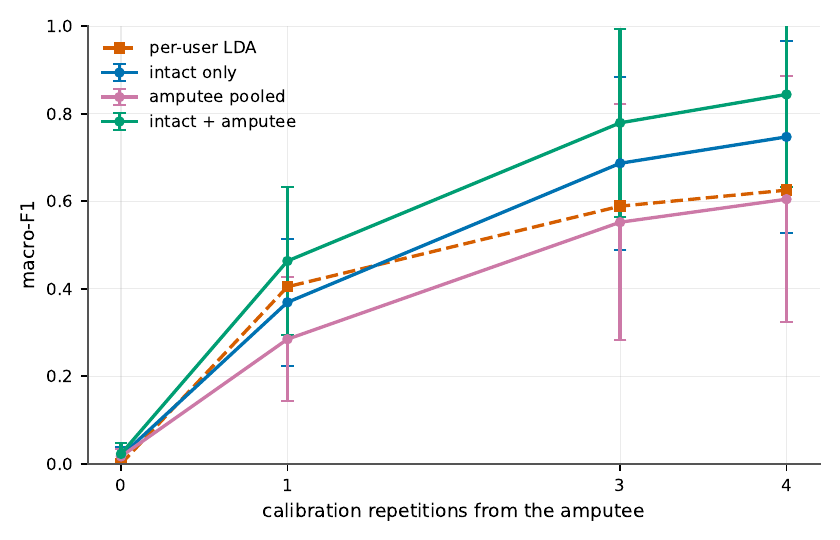}
\caption{}\label{fig:ch6-curves}
\end{subfigure}

\vspace{1.5ex}

\begin{subfigure}[b]{\linewidth}\centering
\includegraphics[width=\linewidth]{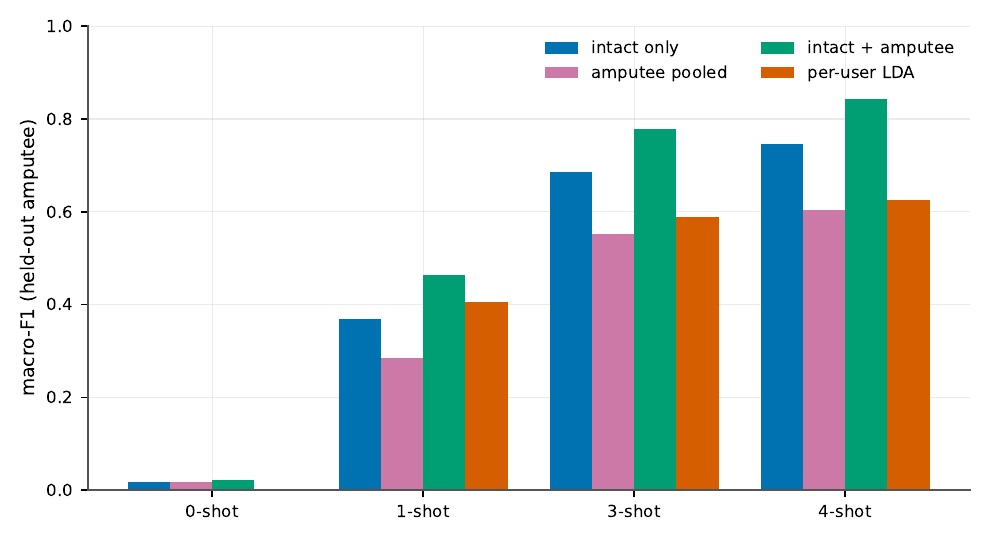}
\caption{}\label{fig:ch6-transfer}
\end{subfigure}
\caption{Transfer against the calibration budget. (a) Zero-shot cross-population transfer. With no
labelled data from the amputee, every training source scores near chance. (b) Calibration curves
on the amputee target. Every training source needs labelled repetitions from the held-out amputee
before it decodes at all. (c) Intact-to-amputee transfer on DB3. Cross-population macro-F1 for
each training source at each calibration budget, with the per-user LDA baseline. Without
calibration the transfer collapses; with a few labelled repetitions the combined source is well
ahead.}
\label{fig:ch6-budget}
\end{figure}

When labeled data from the held-out amputee is unavailable, each training source achieves
nearly-chance performance: 0.018 for the intact-only source, 0.019 for the amputee-pooling
source, and 0.018 for both sources combined (Figure~\ref{fig:ch6-zeroshot}). A decoder capable
of decoding previously unseen intact subjects at 0.198 on the same database family as described
in the companion calibration-budget study~\citep{odeyemi_fewshot} achieved nearly nothing when
decoding an amputee that it has never seen.

Regardless of what knowledge or ability the encoder derived from analyzing forty intact
forearms, it could not, on its own, describe what an amputee's residual limb produces. Thus,
cross-population generalization and cross-user generalization are fundamentally distinct issues.
Success at one issue does not necessarily imply progress at the second issue.

\subsection{Effect of training-source composition}

\begin{figure}[tbp]\centering
\begin{subfigure}[b]{0.51\linewidth}\centering
\includegraphics[width=\linewidth]{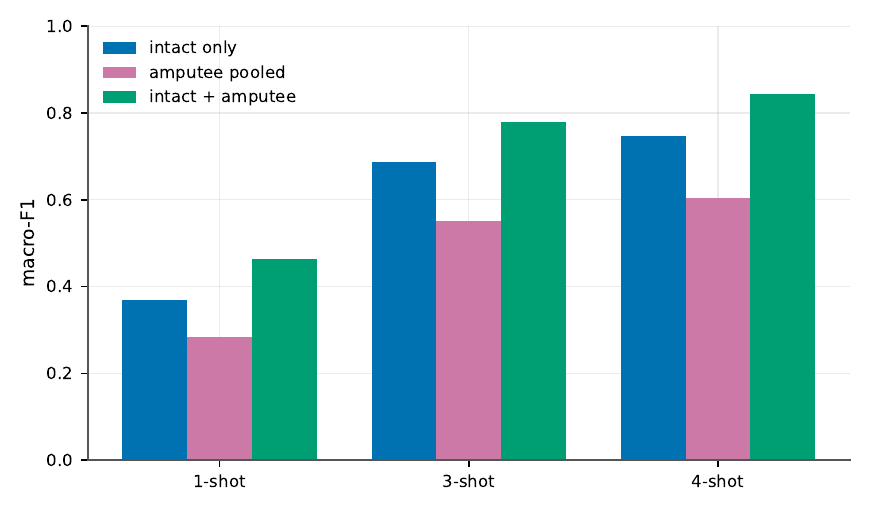}
\caption{}\label{fig:ch6-sources}
\end{subfigure}\hfill
\begin{subfigure}[b]{0.47\linewidth}\centering
\includegraphics[width=\linewidth]{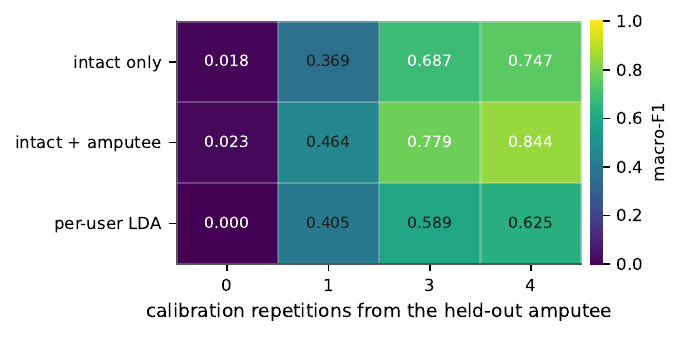}
\caption{}\label{fig:ch6-matrix-heatmap}
\end{subfigure}

\vspace{1.5ex}

\begin{subfigure}[b]{\linewidth}\centering
\includegraphics[width=\linewidth]{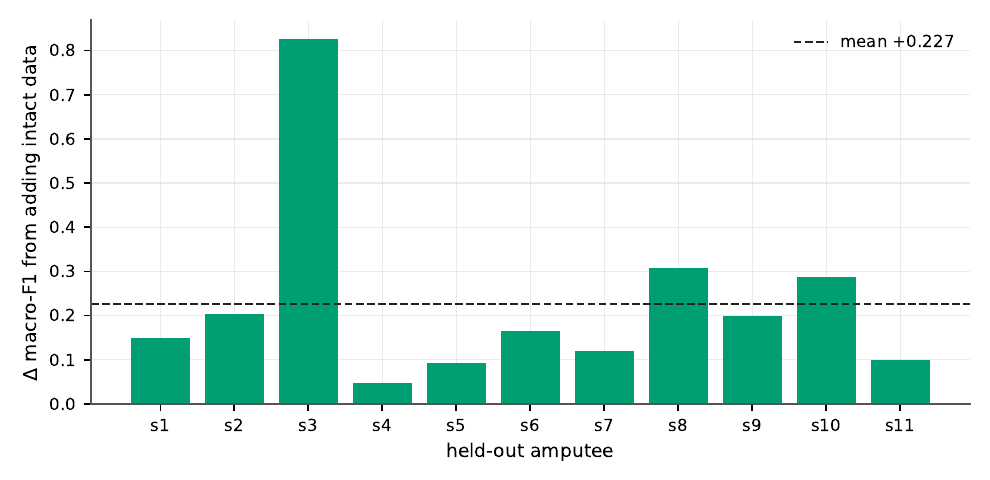}
\caption{}\label{fig:ch6-gain}
\end{subfigure}
\caption{Effect of training-source composition. (a) Transfer by training source. Training on forty
intact subjects transfers to a new amputee better than training on the ten other amputees, and
combining both is better still. (b) Intact-to-amputee transfer matrix: macro-F1 for every training
source at every calibration budget, mean over the eleven amputees. The per-user LDA baseline
cannot be fitted at zero repetitions. (c) Per-amputee benefit of adding forty intact subjects to
the amputee-only training pool.}
\label{fig:ch6-composition}
\end{figure}

Once some labeled data from the target subject becomes available, all three sources decode, and
they separate based on training population (Figures~\ref{fig:ch6-curves}, \ref{fig:ch6-transfer},
\ref{fig:ch6-sources} and~\ref{fig:ch6-matrix-heatmap},
Table~\ref{tbl:ch6-matrix}). Training on forty intact subjects reached 0.687 at three
repetitions. Training on the ten additional amputees who are in the same population as the
target reached 0.552. The intact-only source performed 0.135 higher than the amputee-pooled
source (at $p = 2.9\times10^{-3}$) on ten of the eleven subjects
(Table~\ref{tbl:ch6-significance}).

Using only ten amputees is insufficient to train a decoder model from. Using forty intact
forearms provides enough commonalities about how forearm muscles generate hand motion that they
are valuable regardless of being the wrong population. The comparison made here is between
existing source pools and not between populations of equal sizes. Therefore, pool-size needs to
be considered before population is credited. It can be done, at least approximately: the
companion encoder study~\citep{odeyemi_encoder} measured the impact of subject count directly on
DB2 while holding population constant, and on a similar scale, decreasing the number of training
subjects from thirty-nine down to nine decreased macro-F1 by approximately $0.012$, an
order-of-magnitude smaller than the $0.135$ observed in this study. Therefore, pool-size cannot
explain this difference; thus, this difference is attributable to training population. However,
it also contains a caveat: because a measure of subject count obtained through analysis of intact
subjects may not generalize to an equivalent measure of subject count obtained through analysis
of amputee subjects, it cannot be assumed that this measure will carry-over similarly between
populations.

Therefore, this result allows for a practical interpretation: when recruitment of amputee
subjects constitutes a limiting factor in research design, then a large dataset consisting
solely of intact subjects is the better of the two sources available today.

Combining the amputee subjects with the intact subjects improves performance further; by $0.092$
at three repetitions on all eleven subjects (Figure~\ref{fig:ch6-gain}). Therefore, these two
source pools represent complementary options as opposed to redundant ones.

\begin{table}[t]\centering
\caption{Paired comparisons across the eleven amputees. Positive differences favour the first named condition.}\label{tbl:ch6-significance}
{\footnotesize\setlength{\tabcolsep}{4pt}%
\begin{tabular}{llccc}
\toprule
Comparison & Budget & Mean $\Delta$ & Subjects favouring & Wilcoxon $p$ \\
\midrule
intact+amputee vs LDA & 1-shot & +0.059 & 7/11 & 5.4e-02 \\
intact+amputee vs LDA & 3-shot & +0.190 & 11/11 & 9.8e-04 \\
intact+amputee vs LDA & 4-shot & +0.219 & 11/11 & 9.8e-04 \\
\midrule
intact only vs amputee pooled & 1-shot & +0.084 & 8/11 & 6.7e-02 \\
intact only vs amputee pooled & 3-shot & +0.135 & 10/11 & 2.9e-03 \\
intact only vs amputee pooled & 4-shot & +0.142 & 11/11 & 9.8e-04 \\
\midrule
intact+amputee vs intact only & 1-shot & +0.094 & 10/11 & 2.9e-03 \\
intact+amputee vs intact only & 3-shot & +0.092 & 11/11 & 9.8e-04 \\
intact+amputee vs intact only & 4-shot & +0.097 & 11/11 & 9.8e-04 \\
\bottomrule
\end{tabular}}
\end{table}

Two of the three source-condition analyses demonstrated sensitivity to the optimization method
employed in a manner analogous to that demonstrated by the subject-count sweep in the companion
encoder study~\citep{odeyemi_encoder}. When employing a learning rate of $4\times 10^{-4}$, two
folds representing the amputee-pooling source condition (with its training pool consisting of
only ten subjects) collapsed to a single-class predictive model. At $1.5\times 10^{-4}$
learning-rate neither of those two folds failed to converge; however, one fold represented by an
intact-only source failed to converge. Therefore, every condition was evaluated at both rates;
furthermore, for each subject and condition, the converged run is utilized. In addition to
recording the results associated with each cell, the learning rate employed for each cell is also
recorded. Reporting means based on both converged and collapsed runs collectively in an earlier
version of this analysis would underestimate the amputee-pooling source by approximately 11
points and increase the difference between the two sources correspondingly.

\subsection{Comparison with the per-user baseline}

\begin{table}[t]\centering
\caption{The combined-source encoder against the per-user Hudgins and LDA pipeline on the amputee
target, at matched calibration budget, with the paired outcome across the eleven subjects. The
last three columns give the test of the interaction predicted in \encoderstudyref: because the
per-user baseline is weaker on amputees, the encoder's margin over it should be larger there. The
prediction holds at three repetitions and fails at one.}
\label{tbl:ch6-vslda}\label{tbl:ch6-prediction}
{\scriptsize\setlength{\tabcolsep}{3pt}%
\begin{tabular}{lccccccl}
\toprule
Budget & Encoder & \shortstack{LDA baseline\\(amputee)} & \shortstack{$\Delta$ margin\\(amputee)} &
  \shortstack{Subjects\\ahead} & \shortstack{Baseline\\(intact)} & \shortstack{Margin\\(intact)} &
  Outcome \\
\midrule
1-shot & 0.464 & 0.405 & +0.059 & 7/11 & 0.671 & +0.126 & not confirmed \\
3-shot & 0.779 & 0.589 & +0.190 & 11/11 & 0.857 & +0.110 & confirmed \\
4-shot & 0.844 & 0.625 & +0.219 & 11/11 & -- & -- & -- \\
\bottomrule
\end{tabular}}
\end{table}

\begin{figure}[tbp]\centering
\begin{subfigure}[b]{0.53\linewidth}\centering
\includegraphics[width=\linewidth]{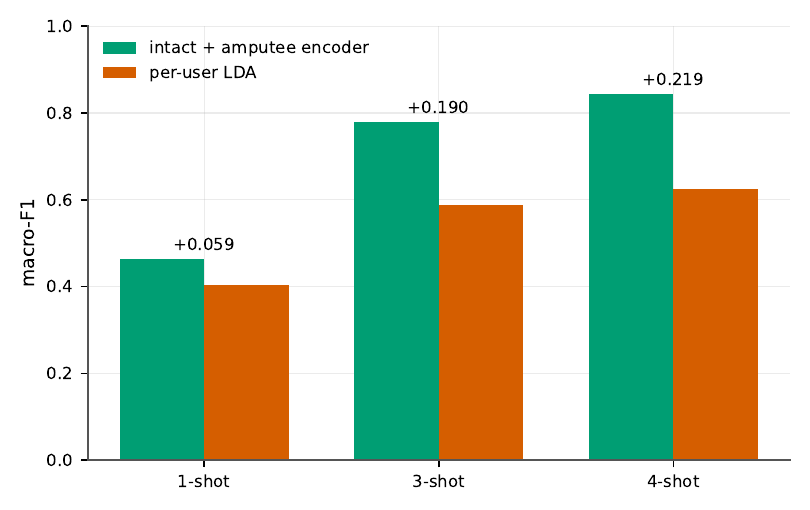}
\caption{}\label{fig:ch6-vslda}
\end{subfigure}\hfill
\begin{subfigure}[b]{0.45\linewidth}\centering
\includegraphics[width=\linewidth]{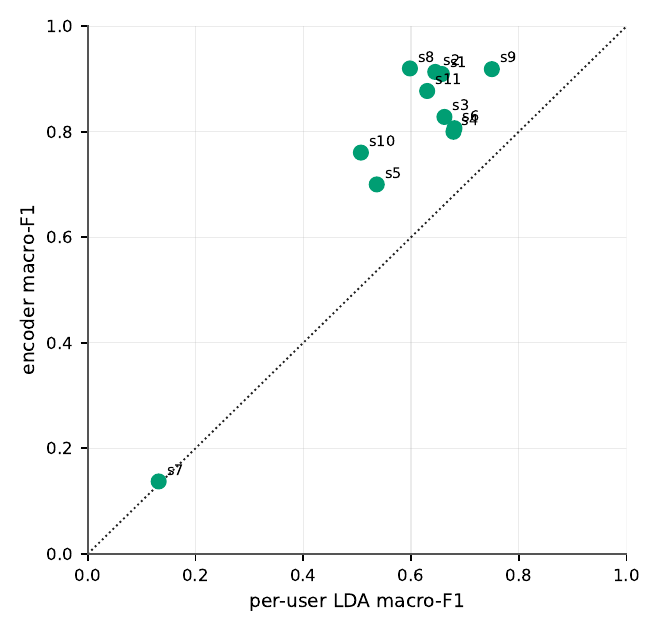}
\caption{}\label{fig:ch6-scatter}
\end{subfigure}

\vspace{1.5ex}

\begin{subfigure}[b]{\linewidth}\centering
\includegraphics[width=\linewidth]{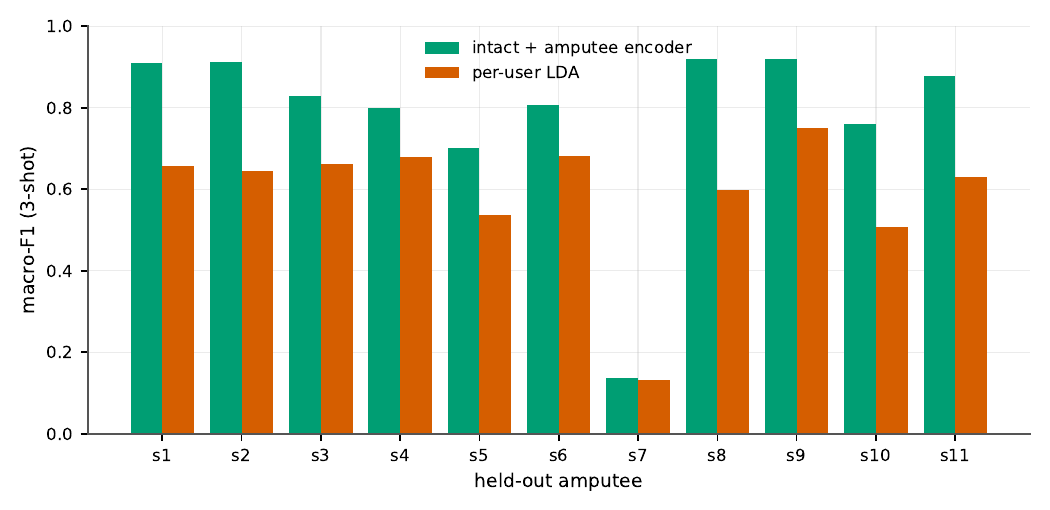}
\caption{}\label{fig:ch6-persubject}
\end{subfigure}
\caption{The encoder against the per-user baseline. (a) The combined-source encoder against the
per-user Hudgins and LDA pipeline fitted on the same repetitions of the same amputee.
(b) Per-amputee encoder score against the per-user LDA baseline at three repetitions. Points above
the diagonal are subjects for whom the encoder is better. (c) Per-amputee macro-F1 at three
calibration repetitions. The combined-source encoder exceeds the per-user LDA baseline for every
one of the eleven subjects.}
\label{fig:ch6-baseline}
\end{figure}

The combined-source encoder performs better at every budget than classifiers constructed
individually for each user based upon the same repetitions of EMG measurements from that user
(Table~\ref{tbl:ch6-vslda}, Figure~\ref{fig:ch6-vslda}). At one repetition, it outperformed
individual classifiers by 0.059, but only for seven out of eleven subjects at $p = 0.054$. Thus,
at that budget level, there is not sufficient evidence that the encoder improves upon individual
user-based classifiers. At three repetitions it outperforms by 0.190 and at four repetitions by
0.219 for all eleven subjects ($p = 9.8 \times 10^{-4}$). The view presented in
Table~\ref{tbl:ch6-persubject} (Figures~\ref{fig:ch6-persubject} and~\ref{fig:ch6-scatter})
demonstrates that all eleven amputees scored above the diagonal line at three repetitions;
therefore, the margin is not carried by a subset of easy subjects.

\subsection{Testing the pre-registered prediction}

The companion encoder study~\citep{odeyemi_encoder} found no single answer among the three
datasets studied. The encoder outperforms the per-user LDA baseline substantially on DB1,
slightly on DB2 after being trained adequately, and falls behind it on DB5. Notably, none of
these outcomes depend on sampling rate or number of channels directly and not even on number of
training subjects over the range available. What tracks that ordering is the strength of the
per-user LDA baseline. As signal quality improves (measured through fidelity), so too does
performance of the per-user LDA baseline: on DB1, it achieves an average macro-F1 value of
$0.593$ at 100~Hz; on DB5, it achieves an average macro-F1 value of $0.802$ at 200~Hz; and on DB2
it achieves an average macro-F1 value of $0.857$ at 2~kHz. Hand-crafted features obviously have
greater bandwidth to capitalize upon as sampling rates increase. Thus, the margins open to
trainable encoders are least when baselines are strongest. Separately, constraints placed by
limiting subject diversity are a stability bound rather than a continuous decline. Down to nine
subjects however, the encoder retains most of its advantages versus the per-user classifier;
below that it ceases training altogether as opposed to degrading progressively.

That study established that whether a cross-user encoder learns sufficiently well to exceed a
per-user classifier is tracked by how strong that per-user classifier is, with the training pool
binding only as a stability floor. The benchmark specification for this paper extended that rule
with a premise it did not itself contain, that EMG from an amputee subject is less separable than
EMG from intact subjects; it hypothesized that this implies that a per-user classifier will be
weaker on amputee subjects than on intact subjects; and finally, it predicted that the encoder's
margin over that baseline will be larger on amputees than on intact subjects at matched budgets.

The premise holds true; i.e., the per-user baseline is weaker on amputees than on intact
subjects: 0.589 vs. 0.857 at three repetitions. Whether or not this conclusion holds depends on
whether the amputee side uses a source pool containing intact subjects only or whether it uses
an additional source pool composed entirely of amputee subjects. The reference value for intact
subjects is 0.110 and was generated by an encoder trained solely on intact data; thus, the
comparison must be made against the intact-only source. Therefore, at three repetitions, the
amputee margin under that source is $+0.092$. Since this margin is less than 0.110, it does not
confirm our prediction. The prediction is confirmed only when combining both source pools;
specifically, adding ten additional amputee subjects lifts the margin to 0.190, i.e., a
same-population supplementation that does not exist for comparisons involving only intact
subjects. At one repetition it fails under either source, the combined margin being 0.058
against an intact margin of 0.126, as shown in Table~\ref{tbl:ch6-prediction} and
Figure~\ref{fig:ch6-prediction}.

Thus, we observe support for interaction only under combined use of both source pools; moreover,
this interaction is bounded by limitations unforeseen by original predictions. Weaknesses in
per-user baselines create larger advantages for encoders; but only once there is sufficient data
from target subjects for encoders to utilize those weaknesses. At a single repetition the
population shift prevails, and encoders are still recovering from it; meanwhile, per-user
classifiers, whatever their ceiling, are at least fitted to their respective target subjects.

\providecommand{\thisdocnoun}{chapter}
\begin{figure}[t]\centering
\includegraphics[width=0.72\textwidth]{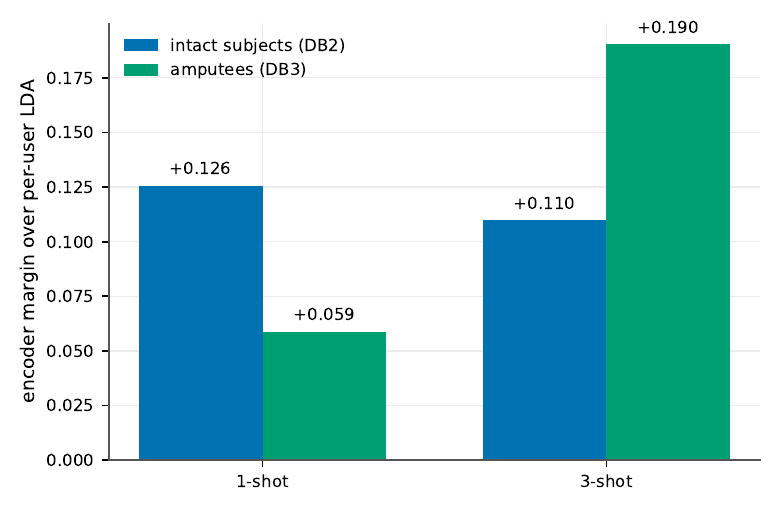}
\caption{Test of the interaction predicted in this \thisdocnoun{}'s pre-registered benchmark specification. The encoder's margin over the per-user LDA baseline is larger on amputees than on intact subjects at three repetitions under the combined source, but not at one.}\label{fig:ch6-prediction}
\end{figure}

\subsection{Variability across the cohort}

\begin{figure}[tbp]\centering
\begin{subfigure}[b]{0.82\linewidth}\centering
\includegraphics[width=\linewidth]{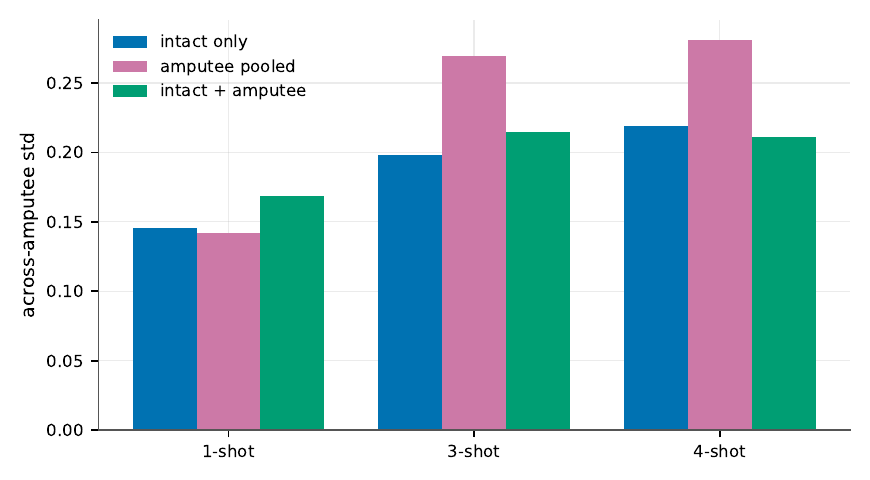}
\caption{}\label{fig:ch6-reliability}
\end{subfigure}

\vspace{1.5ex}

\begin{subfigure}[b]{\linewidth}\centering
\includegraphics[width=\linewidth]{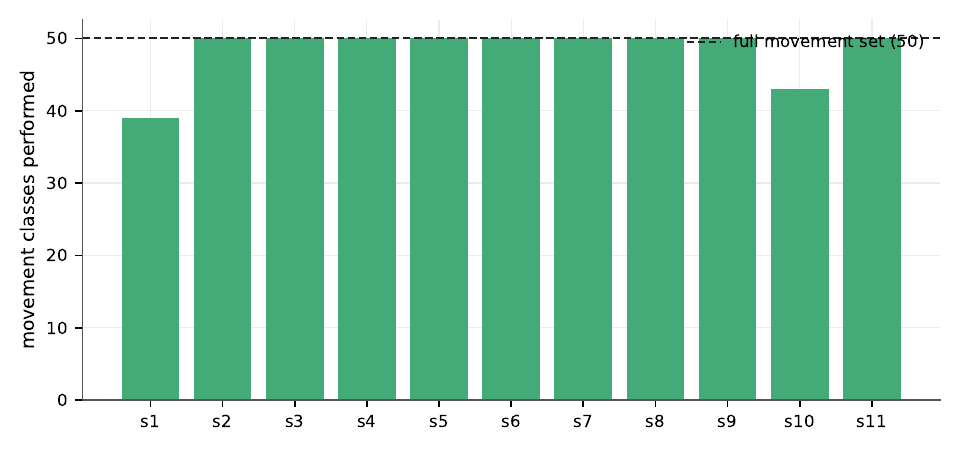}
\caption{}\label{fig:ch6-cohort}
\end{subfigure}
\caption{Variability across the cohort. (a) Across-subject spread of each training source. Amputee
outcomes vary far more than intact-limb outcomes. (b) The amputee cohort. Movement classes
actually performed by each subject; two subjects could not complete the full set.}
\label{fig:ch6-variability}
\end{figure}

Amputee outcomes exhibit far greater variability than outcomes exhibited by subjects with intact
limbs (Figure~\ref{fig:ch6-reliability}, Figure~\ref{fig:ch6-cohort}). Thereby, across-subjects
standard deviation is relatively large compared to mean performance for combined-source
encoding; furthermore, two subjects exhibiting reduced movement sets are not among the two
lowest-performing subjects; thus, variability cannot be explained solely through loss of
movement sets. Subject-specific variables such as stump-length, post-amputation duration and
residual muscular status are documented in database documentation but cannot be resolved into
clean predictors using only eleven subjects.

\section{Discussion}

\providecommand{\encoderstudyshort}{Chapter~3}
\begin{table}[t]\centering
\caption{Summary of what the transfer study establishes.}\label{tbl:ch6-summary}
{\small\setlength{\tabcolsep}{4pt}%
\begin{tabularx}{\textwidth}{@{}>{\raggedright\arraybackslash}X >{\raggedright\arraybackslash}X@{}}
\toprule
Finding & Evidence \\
\midrule
Zero-shot cross-population transfer fails & all sources near chance without calibration \\
Intact data transfers better than scarce amputee data & $+0.135$ at three repetitions \\
Combining both sources is best & $+0.092$ over intact alone \\
The encoder exceeds the per-user baseline on amputees & $+0.190$ at three repetitions, 11/11 subjects \\
The \encoderstudyshort{} interaction holds, with a limit & confirmed at three repetitions, not at one \\
\bottomrule
\end{tabularx}}
\end{table}

A model that was applied to each of the eleven amputees without architectural change exceeded
the per-user pipeline that a clinic would fit, for every subject in the cohort, after just three
labelled repetitions were available (Table~\ref{tbl:ch6-summary}).

Unexpectedly, intact limb data transferred better than amputee data. If you have a choice
between using forty subjects from the wrong population versus ten from the right one, the wrong
population is ahead by a wide margin. Using both populations is even better than using either of
them alone. Thus, manufacturers do not need to wait until they have a substantial corpus of
amputee data to train a decoder, and researchers do not need to discard intact limb data as
irrelevant to the amputee case. Rather than being used as the training set, amputee data should
be viewed as the additive component that adapts a model already built elsewhere.

The results showed that the zero-shot condition failed. Regardless of how many intact limb
trials were completed, no usable decoder was produced for any unseen amputee without labels.
Some form of individualized calibration is needed. The question of how little may be possible is
addressed in the companion calibration-budget study~\citep{odeyemi_fewshot}, where the answer is
one to three repetitions.

There are four limits to how broadly the results apply. There are only eleven subjects in the
cohort. The heterogeneity of the population represented by these subjects is reflected in the
across-subject standard deviations listed in Table~\ref{tbl:ch6-matrix}. The evaluation is
conducted offline and within session, therefore it does not address what happens when an amputee
takes off and puts back on the device, which the companion session
study~\citep{odeyemi_intersession} treats separately on intact subjects. In addition, the
movements are prompted rather than functional, thus recognizing them is different than
controlling a hand while performing a task. Finally, the two sources of data differ in quantity
as well as in terms of population, forty subjects vs. ten; the subject count effect measured on
DB2 is far too small to account for the difference in performance observed between intact limb
data and amputee data, however it is measured on intact limb data and at a single model size,
therefore the population interpretation relies upon extrapolation.

\section{Conclusion}

When carried over to eleven transradial amputees, on a protocol that matches its intact-limb
training data, the encoder requires labeled data from the new user before it begins decoding and
exceeds the per-user classifier a clinic would fit by 0.190 macro F1 at three repetitions and
for every subject in the cohort. Training on forty intact subjects produces better transfers to
a new amputee than training on ten other amputees, and combining the two produces better
transfers than either individually. The prediction pre-registered for this study, which extends
the companion encoder study's rule~\citep{odeyemi_encoder} with the premise that amputee EMG is
less separable, holds true only after a few repetitions become available and after enriching the
source pool with additional amputees. At a single repetition, and at every budget under a source
matched to the intact-limb comparison, it fails. Thus, it locates the boundary of the proposed
account.

\section*{CRediT authorship contribution statement}
\textbf{Jethro Odeyemi:} Conceptualization, Methodology, Software, Formal analysis,
Investigation, Data curation, Visualization, Writing -- original draft, Writing -- review and
editing.

\textbf{W.J. (Chris) Zhang:} Conceptualization, Methodology, Resources, Supervision,
Project administration, Writing -- review and editing.

\section*{Declaration of competing interest}
The authors declare no competing financial interests or personal relationships that could have
appeared to influence the work reported in this paper.

\section*{Funding}
This research did not receive any specific grant from funding agencies in the public, commercial, or not-for-profit sectors.

\section*{Ethics statement}
This study is a secondary analysis of the NinaPro DB2 and DB3 databases, which are de-identified
and publicly released by the original investigators. No new human data were collected for this
work, and no participants were recruited, contacted, or identifiable to the author. Ethical
approval and informed consent for the original recordings were obtained by the NinaPro
consortium and are reported in the source publications for those databases
\citep{atzori2014ninapro}. On that basis the present study did not require separate ethical
approval.

\section*{Data availability}
This study uses only publicly available data. The NinaPro databases are available from the
NinaPro consortium.

\bibliographystyle{elsarticle-num}
\bibliography{refs}

\end{document}